\documentclass[10pt,twocolumn,letterpaper]{article}

\usepackage{cvpr}              

\definecolor{cvprblue}{rgb}{0.21,0.49,0.74}
\usepackage[pagebackref,breaklinks,colorlinks,allcolors=cvprblue]{hyperref}
\usepackage{fontawesome5}
\usepackage[accsupp]{axessibility}
\usepackage[normalem]{ulem}
\usepackage{caption}
\usepackage{graphicx}
\usepackage{booktabs}
\usepackage{enumitem}
\usepackage{booktabs} 
\usepackage{bm}
\usepackage{amsmath}
\usepackage{paralist}
\usepackage{multirow}
\usepackage{amsmath}  
\usepackage[ruled]{algorithm2e} 
\usepackage{xcolor}
\usepackage{amsfonts}
\usepackage{dblfloatfix}
\usepackage{pifont}
\usepackage[T1]{fontenc}
\usepackage{fonttable}

\SetAlFnt{\small}
\SetAlCapFnt{\small}
\SetAlCapNameFnt{\small}
\SetAlCapHSkip{0pt}
\def\paperID{416} 
\def\confName{CVPR}
\def\confYear{2025}

\title{\includegraphics[width=0.6cm]{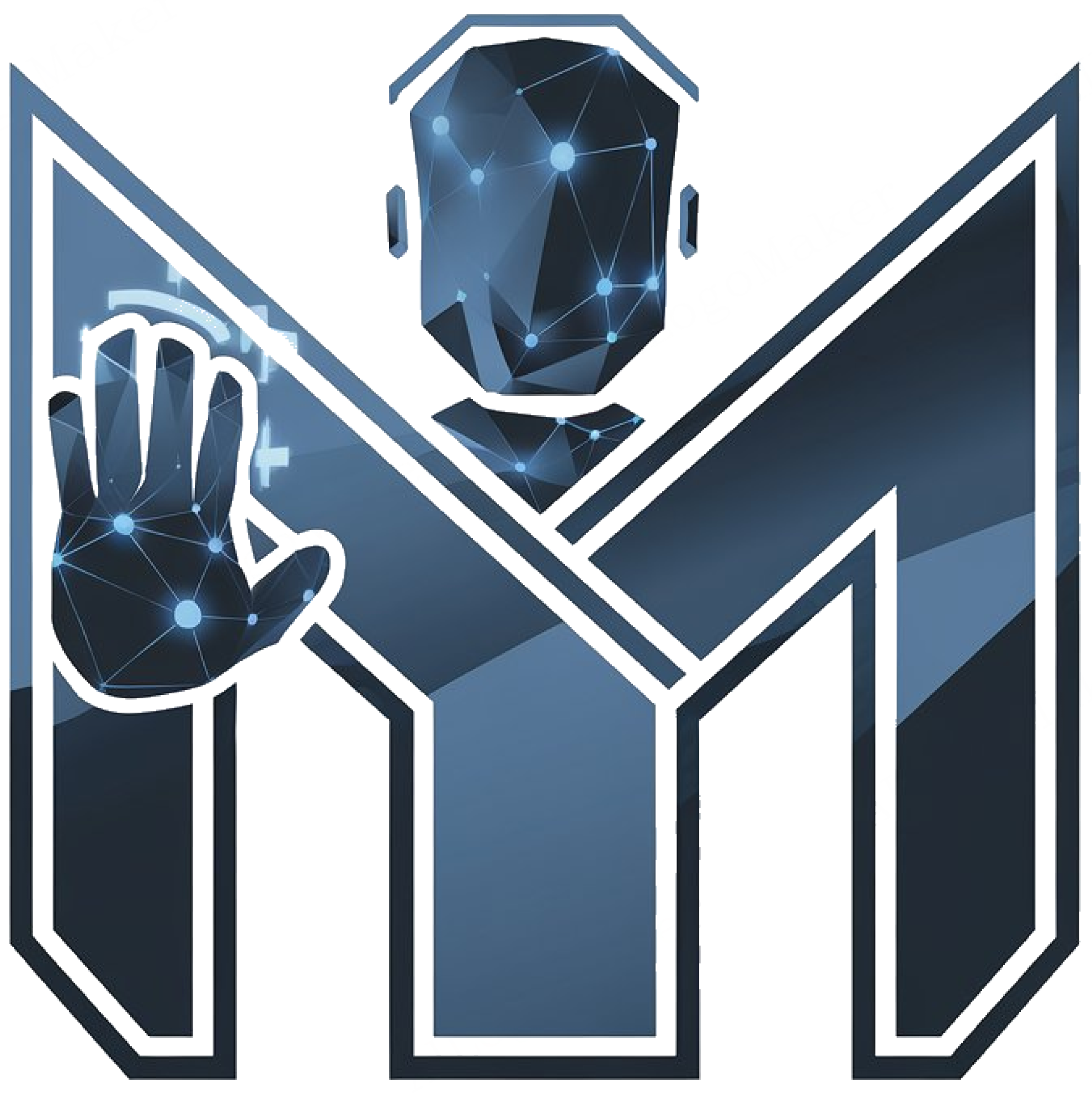} MeGA: Hybrid Mesh-Gaussian Head Avatar for High-Fidelity Rendering and Head Editing}

\author{
Cong Wang$^{1}$, Di Kang$^{2}$, Heyi Sun$^{1}$, Shenhan Qian$^{3}$, Zixuan Wang$^{4}$, \\ Linchao Bao$^{2}$, {\textsuperscript{$\star$}}Song-Hai Zhang$^{1}$ \\ \\
$^{1}$Tsinghua University, $^{2}$Tencent, $^{3}$Technical University of Munich, $^{4}$Carnegie Mellon University \\
\footnotesize{\textsuperscript{$\star$}Corresponding author: shz@tsinghua.edu.cn
\vspace{-2mm}
}
}
\newcommand{\name}{MeGA}

\newcommand{\eqskip}{3pt}
\newcommand{\bfskip}{1mm}

\newcommand{\supp}{\textbf{\textcolor{olive}{Supp. Mat.}}}

\definecolor{ao}{rgb}{0.0, 0.0, 1.0}
\definecolor{airforceblue}{rgb}{0.36, 0.54, 0.66}
\definecolor{ceruleanblue}{rgb}{0.16, 0.32, 0.75}
\definecolor{cerulean}{rgb}{0.0, 0.48, 0.65}
\definecolor{celestialblue}{rgb}{0.29, 0.59, 0.82}
\definecolor{azure(colorwheel)}{rgb}{0.0, 0.5, 1.0}
\definecolor{babyblue}{rgb}{0.54, 0.81, 0.94}
\definecolor{babyblueeyes}{rgb}{0.63, 0.79, 0.95}
\definecolor{ballblue}{rgb}{0.13, 0.67, 0.8}

\definecolor{asparagus}{rgb}{0.53, 0.66, 0.42}
\definecolor{ao(english)}{rgb}{0.0, 0.5, 0.0}
\definecolor{applegreen}{rgb}{0.55, 0.71, 0.0}
\definecolor{armygreen}{rgb}{0.29, 0.33, 0.13}
\definecolor{gray-asparagus}{rgb}{0.27, 0.35, 0.27}
\definecolor{green(ryb)}{rgb}{0.4, 0.69, 0.2}

\definecolor{amethyst}{rgb}{0.6, 0.4, 0.8}
\definecolor{antiquefuchsia}{rgb}{0.57, 0.36, 0.51}
\definecolor{blue-violet}{rgb}{0.54, 0.17, 0.89}
\definecolor{brightlavender}{rgb}{0.75, 0.58, 0.89}
\definecolor{brightube}{rgb}{0.82, 0.62, 0.91}
\definecolor{brilliantlavender}{rgb}{0.96, 0.73, 1.0}

\definecolor{amber}{rgb}{1.0, 0.75, 0.0}
\definecolor{amber(sae/ece)}{rgb}{1.0, 0.49, 0.0}
\definecolor{atomictangerine}{rgb}{1.0, 0.6, 0.4}
\definecolor{burntorange}{rgb}{0.8, 0.33, 0.0}
\definecolor{burntsienna}{rgb}{0.91, 0.45, 0.32}
\definecolor{cadmiumorange}{rgb}{0.93, 0.53, 0.18}
\definecolor{carrotorange}{rgb}{0.93, 0.57, 0.13}
\definecolor{chocolate(web)}{rgb}{0.82, 0.41, 0.12}
\definecolor{chromeyellow}{rgb}{1.0, 0.65, 0.0}
\definecolor{darkorange}{rgb}{1.0, 0.55, 0.0}
\definecolor{darktangerine}{rgb}{1.0, 0.66, 0.07}
\definecolor{deepcarrotorange}{rgb}{0.91, 0.41, 0.17}
\definecolor{deepsaffron}{rgb}{1.0, 0.6, 0.2}
\definecolor{fulvous}{rgb}{0.86, 0.52, 0.0}

\begin{document}
\maketitle

\begin{abstract}
Creating high-fidelity head avatars from multi-view videos is essential for many AR/VR applications.
However, current methods often struggle to achieve high-quality renderings across all head components (e.g., skin vs. hair) due to the limitations of using one \emph{single} representation for elements with varying characteristics.
In this paper, we introduce a Hybrid \uline{\textbf{Me}}sh-\uline{\textbf{G}}aussian Head \uline{\textbf{A}}vatar (\name) that models different head components with more suitable representations. 
Specifically, we employ an enhanced FLAME mesh for the facial representation and predict a UV displacement map to provide per-vertex offsets for improved personalized geometric details.
To achieve photorealistic rendering, we use deferred neural rendering to obtain facial colors and decompose neural textures into three meaningful parts.
For hair modeling, we first build a static canonical hair using 3D Gaussian Splatting.
A rigid transformation and an MLP-based deformation field are further applied to handle complex dynamic expressions.
Combined with our occlusion-aware blending, {\name} generates higher-fidelity renderings for the whole head and naturally supports diverse downstream tasks.
Experiments on the NeRSemble dataset validate the effectiveness of our designs, outperforming previous state-of-the-art methods and enabling versatile editing capabilities, including hairstyle alteration and texture editing.
The code is released in 
{\hypersetup{urlcolor=olive}
\href{https://github.com/conallwang/MeGA}{\faIcon{globe}~\textbf{our github repo}}}.
\end{abstract}

\vspace{-1mm}
\section{Introduction}
\label{sec:intro}

Generating photorealistic rendering of animatable head avatars has been a long-standing focus in computer vision and graphics, with applications spanning AR/VR communication~\cite{DBLP:conf/ismar/HeDP20, DBLP:conf/uist/Orts-EscolanoRF16, DBLP:journals/tog/LombardiSSS18}, gaming~\cite{waggoner2009my}, and remote collaborations~\cite{DBLP:journals/rcim/WangBBZZWHYJ21}.

Existing methods have explored mesh-based representations~\cite{DBLP:journals/tog/LombardiSSS18, DBLP:conf/cvpr/MaSSWLTS21, DBLP:conf/cvpr/GrassalPLRNT22, DBLP:conf/cvpr/BaiTHSTQMDDOPTB23}, NeRF-based representations~\cite{DBLP:conf/siggraph/XuWZ0L23, DBLP:conf/siggrapha/WangKCBSZ23, DBLP:conf/cvpr/0003PXLZ22, DBLP:journals/tog/LombardiSSZSS21}, 3D Gaussians-based representations~\cite{DBLP:conf/cvpr/QianKS0GN24, DBLP:conf/eccv/DhamoNMSSZP24, DBLP:conf/cvpr/XiangGGZ24, wang2024gaussianhead} and achieved remarkable progress in this field.
However, the human head is a complex ``object'' containing components with drastically different characteristics so there may not exist one \emph{single} representation that can model all of them well simultaneously.
For instance, the human hair contains volumetric thin structures while the human face is predominantly surface-like regions and can be animated in a low dimensional space~\cite{DBLP:journals/tog/LiBBL017}.
Thus, using only one representation to model different head components inevitably sacrifices the rendering quality of one part for another.

Ideally, we expect the head avatar representation can be rendered in photorealistic quality and can be easily controlled to perform vivid facial animations.
For high-quality \emph{facial} rendering and animation, Pixel Codec Avatars (PiCA)~\cite{DBLP:conf/cvpr/MaSSWLTS21}, which adopts neural texture representation~\cite{DBLP:journals/tog/ThiesZN19}, have demonstrated extraordinary rendering quality and subtle dynamic texture details while being able to be animated easily due to its mesh-based representation.
However, it contains noticeable artifacts including texture-like hair rendering and mesh-like hair boundaries.
In contrast, GaussianAvatars~\cite{DBLP:conf/cvpr/QianKS0GN24}, which adopts rigged 3D Gaussian Splatting (3DGS)~\cite{DBLP:journals/tog/KerblKLD23} representation, successfully reconstructs high-frequency volumetric human hair but shows inferior facial texture details (e.g., wrinkles) and interpenetration artifacts (e.g., Fig.~\ref{fig:sota}, first row).
Additionally, anti-aliasing of 3DGS remains an open problem~\cite{DBLP:journals/corr/abs-2403-19615, DBLP:conf/cvpr/YuCHS024, DBLP:journals/corr/abs-2408-06286}, significantly impairing its rendering quality on human faces, particularly when zooming in/out.

Therefore, we propose to use more suitable representations for different head components (i.e., neural mesh for the face and 3DGS for the hair), resulting in a Hybrid \textbf{Me}sh-\textbf{G}aussian Head \textbf{A}vatar (\name).
Specifically, we adopt the FLAME mesh~\cite{DBLP:journals/tog/LiBBL017} as our base mesh to model dynamic human faces.
Additionally, we learn a UV displacement map conditioned on the driving signal (i.e., FLAME parameters) to account for the geometric details that cannot be represented in the FLAME space.
For photorealistic rendering, we use neural texture and deferred neural rendering techniques~\cite{DBLP:journals/tog/ThiesZN19,DBLP:conf/cvpr/MaSSWLTS21}.
Unlike PiCA~\cite{DBLP:conf/cvpr/MaSSWLTS21}, our neural texture consists of three components, 
including a diffuse texture map to model the base color, an \emph{expression-dependent} texture map to model dynamic textures (e.g., wrinkles and dimples), 
and a \emph{view-dependent} texture map to handle view-dependent effects.
For hair modeling, we build an canonical 3DGS hair from a chosen frame, which is subsequently deformed by a rigid transformation and an MLP network to capture dynamic hair motion.

Another crucial component of {\name} for high-quality head renderings is the occlusion-aware blending for face and hair images.
Specifically, we conduct occlusion test using our ``near-z'' GS depths rather than commonly used integrated GS depth, enabling more stable training.
To minimizing blending artifacts, we propose an early-stopping strategy during the GS hair rendering to exclude the occluded Gaussians,
combined with a soft-blending technique to create smoother blending boundaries (e.g., hairline).

With this decomposed representation, {\name} not only achieves state-of-the-art rendering quality for the complete head but also supports a range of downstream operations, including hairstyle alterations and texture editing.

Our contributions are summarized below:
\begin{compactitem}
    \item We are the first to propose a hybrid mesh-Gaussian full-head representation, adopting more suitable representations to model different head components (i.e., neural mesh for the face, 3DGS for the hair).
    \item The decomposed representation naturally supports various downstream applications, including high-quality hair alteration and texture editing.
    \item Experimental results on the NeRSemble dataset show that our approach produces higher-quality renderings for novel expressions and views.
\end{compactitem}

\vspace{-1mm}
\section{Related Works}

\subsection{Animatable Head Avatars}

Creating high-fidelity, animatable 3D head avatars from images or videos has always been of great interest in the computer vision and graphics community. Traditional explicit geometric modeling methods \cite{ichim2015dynamic,hu2017avatar,bao2021high} usually rely on low-poly meshes and suffer from inaccurate details, especially around hair regions. 
With the rise of neural network-based approaches, Codec Avatars \cite{DBLP:journals/tog/LombardiSSS18,DBLP:conf/cvpr/MaSSWLTS21,lombardi2021mixture,wang2021learning,DBLP:conf/siggrapha/WangKCBSZ23, DBLP:conf/cvpr/SaitoSSLN24} utilize coarse tracked meshes together with neural networks to model and render facial performance sequences by capturing them from multi-view videos. 
The captured avatars can be animated using a driving model \cite{DBLP:journals/tog/LombardiSSS18} that maps control signals to the avatar latent codes; however, this approach may lack intuitive controls. 
Another line of work \cite{gafni2021dynamic,DBLP:conf/cvpr/GrassalPLRNT22,hong2022headnerf,zheng2022avatar,gao2022reconstructing,DBLP:conf/cvpr/ZhengYWBH23,xu2023avatarmav,zhao2023havatar,zielonka2023instant,DBLP:conf/cvpr/QianKS0GN24,DBLP:conf/cvpr/XiangGGZ24, DBLP:conf/iros/WangZWDQM21, DBLP:conf/siggrapha/WangLBXZW19} aims to model head avatars that can be directly driven using parameters from existing parametric models (e.g., FLAME~\cite{DBLP:journals/tog/LiBBL017}). 
It is noteworthy that methods utilizing multi-view video inputs \cite{DBLP:conf/cvpr/QianKS0GN24,DBLP:conf/cvpr/XiangGGZ24} typically significantly outperform those relying on monocular inputs \cite{gafni2021dynamic,DBLP:conf/cvpr/GrassalPLRNT22,hong2022headnerf,zheng2022avatar,gao2022reconstructing,DBLP:conf/cvpr/ZhengYWBH23,xu2023avatarmav,zhao2023havatar,zielonka2023instant}. 
Our work follows the multi-view video setting like the GaussianAvatars \cite{DBLP:conf/cvpr/QianKS0GN24}.

\subsection{3D Representations for Head Avatars}

Traditional 3D head avatars \cite{ichim2015dynamic,hu2017avatar,bao2021high} typically employ a topological consistent, morphable mesh model (e.g., 3DMM) \cite{DBLP:conf/siggraph/BlanzV99,DBLP:journals/tog/LiBBL017} for facial modeling and animation. 
However, it is exceedingly challenging to faithfully reconstruct the intricate details of the face and complicated hair regions using standard 3DMMs. 
To address these challenges, implicit head avatar models integrate neural networks into the avatar modeling and rendering processes. 
For instance, the Neural Head Avatar \cite{DBLP:conf/cvpr/GrassalPLRNT22} and IM Avatar \cite{zheng2022avatar} leverage neural networks to model the geometric and texture details beyond the FLAME model \cite{DBLP:journals/tog/LiBBL017}. 
The Deferred Neural Rendering \cite{DBLP:journals/tog/ThiesZN19} approach achieves high-quality, photorealistic rendering with imperfect 3D assets by substituting the graphics rendering pipeline with a neural network-based rendering process. 
In addition to the mesh-based representations~\cite{DBLP:journals/tog/BharadwajZHBA23, DBLP:conf/cvpr/GrassalPLRNT22, zheng2022avatar, DBLP:conf/iccv/WangWM23, DBLP:conf/icra/WangWM22}, there are research works based on point-based representations \cite{DBLP:conf/cvpr/ZhengYWBH23,DBLP:conf/siggrapha/WangKCBSZ23}, volume-based representations \cite{lombardi2019neural,xu2023avatarmav}, the mixture of volumetric primitives \cite{lombardi2021mixture}, NeRF-based representations \cite{gafni2021dynamic,hong2022headnerf,gao2022reconstructing,zhao2023havatar, DBLP:conf/cvpr/ChenOB023}, and more recent 3D Gaussians-based representations \cite{DBLP:conf/cvpr/QianKS0GN24, DBLP:conf/cvpr/XuCL00ZL24,DBLP:conf/eccv/DhamoNMSSZP24, wang2024gaussianhead, DBLP:conf/cvpr/XiangGGZ24, DBLP:conf/siggraph/MaWS024, DBLP:journals/corr/abs-2508-09597}. 
Different from previous methods, we employ a hybrid mesh-Gaussian representation to decouple the modeling of the human face and hair.

Note that GaussianAvatars~\cite{DBLP:conf/cvpr/QianKS0GN24} only uses the mesh as the underlying deformation proxy to obtain an animatable 3DGS-based head.
The potential artifacts (e.g., inferior facial details and interpenetration artifacts) of 3DGS are enlarged due to large scale (e.g., jaw open) and non-rigid deformation (e.g., extreme expressions).
DELTA~\cite{DBLP:journals/corr/abs-2309-06441} leverages mesh and NeRF to model faces and hair separately, which is conceptually similar to our approach.
However, by incorporating deferred neural rendering and advancing from NeRF to 3DGS, our MeGA achieves higher-quality renderings and greatly improved efficiency, while also supporting a broader range of downstream applications.

\begin{figure*}[t]
\centering
\vspace{-1mm}
\includegraphics[width=\linewidth]{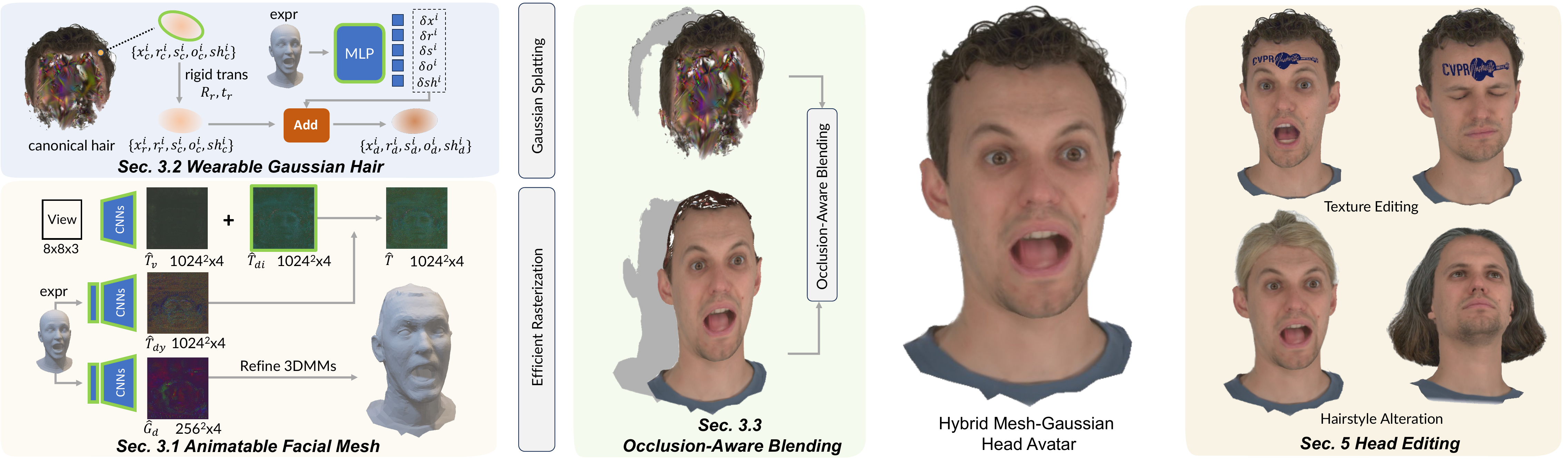}
\caption{\textbf{Hybrid \uline{Me}sh-\uline{G}aussian Head \uline{A}vatar.} 
\name{} models different head components with more suitable representations.
For \emph{facial} modeling, we propose a neural mesh-based representation, including a UV displacement map $\hat{\bm{G}}_d$ for geometric details, a disentangled neural texture map composed by $\hat{\bm{T}}_{di}$, $\hat{\bm{T}}_{dy}$, and $\hat{\bm{T}}_v$ to learn the diffuse colors, dynamic textures, and view-dependent colors, respectively.
For \emph{hair} modeling, a canonical 3D Gaussian Splatting is reconstructed and then animated using a rigid transformation and an MLP-based non-rigid deformation field.
A mesh occlusion-aware blending is proposed to properly blend the face and hair images. 
{\name} naturally supports hair alteration and texture editing due to the disentangled representations.
Learnable parameters are highlighted
using green boxes.
}
\label{fig:overview}
\vspace{-1.5mm}
\end{figure*}

\vspace{-1mm}
\section{Hybrid Mesh-Gaussian Head Avatar}
\label{sec:representation}

%
Our goal is to create an animatable head avatar from multi-view videos that can be driven by FLAME parameters.
Specifically, As illustrated in Fig.~\ref{fig:overview},
%
given the driving signal (i.e., FLAME shape $\beta$, expression $\psi$, and pose $\phi$ parameters)
and view vector $\bm{d}$, we employ three decoders to generate a UV displacement map $\hat{\bm{G}}_d$, a view texture map $\hat{\bm{T}}_v$, and a dynamic texture map $\hat{\bm{T}}_{dy}$.
%
The UV displacement map $\hat{\bm{G}}_d$ captures geometric details beyond the FLAME.
%
The view texture map $\hat{\bm{T}}_v$, dynamic texture map $\hat{\bm{T}}_{dy}$ and diffuse texture map $\hat{\bm{T}}_{di}$ are combined to produce facial neural textures $\hat{\bm{T}}$.
Facial colors are then obtained through efficient mesh rasterization, followed by a lightweight per-pixel decoder.
%
For hair modeling, we create a static canonical 3DGS hair from a chosen frame and incorporate a global rigid transformation and an MLP-based non-rigid deformation field for animation.
%
Finally, a mesh occlusion-aware blending is proposed to properly blend the face and hair images.

%

\vspace{-1mm}
\subsection{Animatable Facial Mesh}

To precisely control head avatars and robust generalization to unseen expressions, we use an enhanced FLAME mesh as our facial geometry, along with a UV displacement map to capture personalized geometric details.
Disentangled neural textures are mapped onto this refined facial mesh and decoded into RGB colors via our per-pixel texture decoder.

\vspace{\bfskip}
\noindent
\textbf{Enhanced FLAME Mesh.}
To increase the expressiveness of FLAME mesh, similar to~\cite{DBLP:conf/cvpr/GrassalPLRNT22}, we densify the FLAME mesh using four-way subdivision and add faces for human teeth, generating our enhanced FLAME mesh:
\begin{equation}
\setlength{\abovedisplayskip}{\eqskip}
\setlength{\belowdisplayskip}{\eqskip}
    \mathcal{T}(\beta, \psi, \phi) = \{\mathcal{V}(\beta, \psi, \phi), \mathcal{F}\},
\end{equation}
where $\mathcal{V} \in \mathbb{R}^{16428 \times 3}$ represents the vertices of the enhanced mesh, calculated using the shape $\beta \in \mathbb{R}^{300}$, expression $\psi \in \mathbb{R}^{100}$, and pose $\phi \in \mathbb{R}^{15}$ parameters via linear blend skinning (LBS).
The faces of the enhanced mesh are denoted by $\mathcal{F} \in \mathbb{R}^{40212 \times 3}$.

%
\vspace{\bfskip}
\noindent
\textbf{Geometry Refinement.}
Building on the enhanced FLAME mesh, and inspired by~\cite{DBLP:conf/siggrapha/WangKCBSZ23, DBLP:journals/corr/abs-2312-14140}, we predict a UV displacement map $\hat{\bm{G}}_d$ conditioned on the FLAME expression parameters $\psi$ and pose parameters $\phi$. The refined mesh $\mathcal{T}_r$ is defined as follows:
\begin{equation}
\setlength{\abovedisplayskip}{\eqskip}
\setlength{\belowdisplayskip}{\eqskip}
\begin{split}
    \mathcal{T}_r(\beta, \psi, \phi) =& \{\mathcal{V}_r(\beta, \psi, \phi), \mathcal{F}\}, \\
    \text{where } \mathcal{V}_r(\beta, \psi, \phi) &= \mathcal{V}(\beta, \psi, \phi) + \mathcal{S}(\hat{G}_d).
\end{split}
\end{equation}
$\mathcal{S}(\cdot)$ samples values based on the UV coordinates.

In contrast to previous geometry refinement networks \cite{DBLP:conf/cvpr/GrassalPLRNT22} that rely on MLPs to predict per-vertex offsets, our approach uses a UV displacement map, which inherently promotes smoothness in the refined mesh due to the locality properties of Convolutional Neural Networks (CNNs).
%
Additionally, by using $\mathcal{S}(\cdot)$, our geometry refinement supports unlimited mesh resolution, i.e., the computation cost does not increase as the number of vertices increases.

%
\vspace{\bfskip}
\noindent
\textbf{Disentangled Neural Texture.}
Given the strengths of neural textures in expressing high-quality dynamic textures and rendering efficiency~\cite{DBLP:conf/cvpr/MaSSWLTS21}, 
we adopt deferred neural rendering~\cite{DBLP:journals/tog/ThiesZN19} to generate colors for facial regions.
To model observations more reasonably, we disentangle neural textures $\hat{\bm{T}} \in \mathbb{R}^{1024 \times 1024 \times 4}$ into three components:
\begin{equation}
\setlength{\abovedisplayskip}{\eqskip}
\setlength{\belowdisplayskip}{\eqskip}
    \hat{\bm{T}} = \hat{\bm{T}}_{di} + \hat{\bm{T}}_v + \hat{\bm{T}}_{dy}, \text{where } \hat{\bm{T}}_{\text{[ ]}} \in \mathbb{R}^{1024 \times 1024 \times 4}.
\end{equation}
The diffuse texture $\hat{\bm{T}}_{di}$ is defined as learnable parameters, representing the base diffuse colors of each face.
%
The view texture $\hat{\bm{T}}_v$ and dynamic texture $\hat{\bm{T}}_{dy}$ are predicted using CNNs conditioned on the view vector $\bm{d}$ and FLAME expression parameters $\psi$ respectively to capture view-dependent effects and dynamic texture details.

%
\vspace{\bfskip}
\noindent
\textbf{Per-Pixel Texture Decoding.} 
To achieve fast and high-fidelity rendering, we utilize a compact MLP with just 307 learnable parameters for per-pixel decoding~\cite{DBLP:conf/cvpr/MaSSWLTS21} to produce RGB colors.
Unlike PiCA~\cite{DBLP:conf/cvpr/MaSSWLTS21}, our RGB colors are predicted solely from UV coordinates and neural textures, which enhances generalization to unseen expressions.
%
Excluding the XYZ coordinate inputs from the decoder prevents from overfitting to a specific coordinate system, thereby improving the renderings for novel expressions.
%

\subsection{Wearable Gaussian Hair}
We adopt 3DGS~\cite{DBLP:journals/tog/KerblKLD23} for hair modeling since it can better reconstruct high-frequency volumetric structures than mesh-based representations~\cite{DBLP:conf/cvpr/MaSSWLTS21, DBLP:conf/cvpr/GrassalPLRNT22}.
%
Specifically, we first select one training frame (all views) to build a 3DGS-based canonical human hair with static modeling.
For dynamic modeling, a rigid transformation is computed via the ICP algorithm~\cite{DBLP:journals/pami/BeslM92} to align the canonical hair with each new frame.
Additionally, an MLP-based deformation field~\cite{DBLP:conf/siggraph/Chen0LXZYL24, DBLP:conf/cvpr/QianWM0024, DBLP:conf/cvpr/ZhouHXQ24} accounts for subtle non-rigid movements.

\vspace{\bfskip}
\noindent
\textbf{Preliminaries: 3D Gaussian Splatting.}
Given calibrated multi-view images and an initial point cloud (e.g., from SfM~\cite{DBLP:conf/cvpr/SchonbergerF16}), a \emph{static} scene can be reconstructed using a set of anisotropic Gaussians $\mathcal{G} = \{x^i, r^i, s^i, o^i, sh^i\}_{i=1:N}$~\cite{DBLP:journals/tog/KerblKLD23}. Here, $i$ represents the $i$-th Gaussian, 
$N$ the number of Gaussians,
$x^i \in \mathbb{R}^3$ the center of the $i$-th Gaussian,
$r^i \in \mathbb{R}^4$ the orientation (represented by a unit quaternion), 
$s^i \in \mathbb{R}^3$ the scale,
$o^i \in \mathbb{R}$ the opacity,
and $sh^i \in \mathbb{R}^{48}$ the spherical harmonics coefficients (up to degree $3$), used to model view-dependent appearance.

To render a pixel's color $\bm{C}$, all 3d Gaussians intersected with its view vector $\bm{d}$ are blended using alpha blending:
\begin{equation}
\setlength{\abovedisplayskip}{\eqskip}
\setlength{\belowdisplayskip}{\eqskip}
    \bm{C} = \sum_{i} c_i \alpha'_i \prod^{i-1}_{j=1} (1 - \alpha'_j),
\end{equation}
where $c_i$ is the color of the $i$-th Gaussian computed from $sh^i$ and the view vector $\bm{d}$.
The blending weight $\alpha'_i$ is given by evaluating the 2D projection of the $i$-th Gaussian~\cite{DBLP:conf/visualization/ZwickerPBG01} multiplied by $o^i$.
All Gaussians are sorted by depth before performing the alpha blending calculation.

\vspace{\bfskip}
\noindent
\textbf{Static Modeling of the Canonical Hair.}
\label{sec:canonical_hair}
To obtain the canonical human hair $\mathcal{G}_c = \{x^i_c, r^i_c, s^i_c, o^i_c, sh^i_c \}_{i=1:N}$, we optimize a 3DGS from multi-view images of one chosen frame.
%
Note that we initialize the point cloud by sampling on- and off-surface points according to the scalp region of the tracked FLAME mesh and only use image pixels under the hair mask regions for photometric training.


%

\vspace{\bfskip}
\noindent
\textbf{Rigid Hair Transformation between Two Frames.}
To handle head movement between different frames, we compute per-frame rigid transformations $\{R_i, t_i\}_{i=1:N_f}$ relative to the FLAME mesh in the canonical frame using the ICP algorithm~\cite{DBLP:journals/pami/BeslM92}:
{
\small
\begin{equation}
\setlength{\abovedisplayskip}{\eqskip}
\setlength{\belowdisplayskip}{\eqskip}
    (R_i, t_i) = \text{ICP} (\mathcal{V}^{scalp}(\beta_i, \psi_i, \phi_i), \mathcal{V}^{scalp}(\beta_c, \psi_c, \phi_c)),
\end{equation}
}
where $N_f$ represents the total number of training frames,
$\beta_c, \psi_c, \text{and } \phi_c$ the FLAME parameters of the canonical frame,
$\mathcal{V}^{scalp}$ the pre-defined scalp vertices.
ICP$(\cdot)$ computes an alignment (i.e., a rigid transformation) by minimizing the Euclidean distance between the two point sets.
%
%

With the rigid transformations, we obtain initial transformed hair Gaussians $\mathcal{G}_{r} = \{x^i_r, r^i_r, s^i_c, o^i_c, sh^i_c \}_{i=1:N}$ which are used for the next dynamic hair modeling.


\vspace{\bfskip}
\noindent
\textbf{Non-Rigid Hair Deformation between Two Frames.}
To account for variations caused by different poses/expressions and achieve sharper renderings, we learn a non-rigid deformation field parameterized by an MLP $\mathcal{M}_d$:
\begin{equation}
\setlength{\abovedisplayskip}{\eqskip}
\setlength{\belowdisplayskip}{\eqskip}
    \mathcal{M}_d: \psi \rightarrow (\delta x, \delta r, \delta s, \delta o, \delta sh), 
\end{equation}
where $\psi$ represents the FLAME expression parameters.
The final Gaussian hair including both rigid and non-rigid deformations is $\mathcal{G}_d = \{x^i_r + \delta x^i, r^i_r + \delta r^i, s^i_c + \delta s^i, o^i_c + \delta o^i, sh^i_c + \delta sh^i \}_{i=1:N}$.

\subsection{Occlusion-Aware Blending}

\begin{figure*}[t]
\centering
\includegraphics[width=\linewidth]{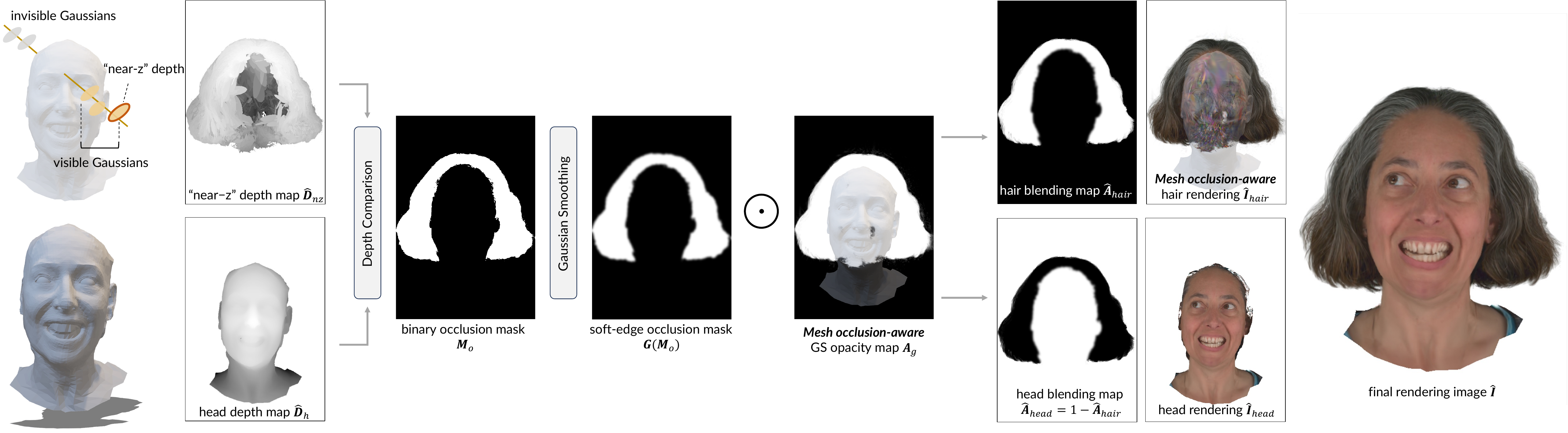}
\caption{\textbf{Mesh Occlusion-Aware Blending.}
By comparing the hair ``near-z'' depth map $\hat{\bm{D}}_{nz}$ and the head depth map $\hat{\bm{D}}_{h}$, we find pixels that should use hair renderings (white regions in $\bm{M}_o$).
Further combining soft-edge occlusion mask $G(\bm{M}_o)$ with \emph{mesh occlusion-aware} hair opacity map $\bm{A}_g$ which only account for visible Gaussians (i.e., in front of the mesh), we obtain the blending map for final renderings.
}
\label{fig:blending}
\end{figure*}

A basic idea for blending is to compare the depth maps of the 3DGS hair and facial mesh, 
and set the color of the final image pixel to that of the closer one (i.e., hard-blending).
In practice, our occlusion-aware blending module (Fig.~\ref{fig:blending}) needs to solve two critical challenges: training stability and blending artifacts (e.g., hairline seams).


\vspace{\bfskip}
\noindent
\textbf{Ensuring Stable Training.} We adopt a simpler but more robust ``near-z'' depth $\hat{\bm{D}}_{nz}$ for our occlusion test, which is defined as the depth value of the first Gaussian (depth sorted) whose opacity value is larger than a predefined threshold (0.05 in our settings).
If an image pixel's ``near-z'' depth is larger than its mesh depth, we know the GS hair is occluded by the facial mesh with high confidence.
In contrast, using 3DGS-rendered depth for occlusion test is unstable because the rendered GS hair depth is close to mesh depth, which fluctuates due to minor training errors and causing a frequently changing occlusion state.
We denote the resulting binary occlusion mask as $\bm{M}_o = \hat{\bm{D}}_{nz} < \hat{\bm{D}}_h$.

\vspace{\bfskip}
\noindent
\textbf{Reducing Blending Artifacts.}
%
Firstly, we propose an early-stopping rendering strategy.
%
Specifically, for regions under $\bm{M}_o$, there exist Gaussians before the mesh, which should be accounted for during rendering, and Gaussians occluded by the mesh, which should be ignored during rendering.
Thus, during the Gaussian rendering process, we will stop the color/alpha accumulation if the next Gaussian (depth sorted) is too far (i.e., the other side of the head) from the current one for a given ray, obtaining an accumulated alpha map $\bm{A}_{g}$ of the Gaussian hair for later blending.
%
%

Then, to further reduce artifacts around blending boundaries (especially the hairline)
in the final renderings, we apply the Gaussian smoothing~\cite{DBLP:journals/corr/abs-2007-09539} to the binary occlusion mask $\bm{M}_o$, resulting in a soft-edge occlusion mask $G(\bm{M}_o)$.

The final blending map for the hair is computed as $\bm{\hat{A}}_{hair} = \bm{A}_{g} \cdot G(\bm{M}_o)$, and the final rendering $\hat{\bm{I}}$ of our hybrid representation is then given by:
\begin{equation}
\setlength{\abovedisplayskip}{\eqskip}
\setlength{\belowdisplayskip}{\eqskip}
    \hat{\bm{I}} = \bm{\hat{A}}_{hair} \cdot \hat{\bm{I}}_{hair} + (\bm{1} - \bm{\hat{A}}_{hair}) \cdot \hat{\bm{I}}_{head}.
\end{equation}
%


\section{Optimizing Head Avatars}
\label{sec:opt}

%
Directly optimizing a complete hybrid facial mesh and Gaussian hair avatar from scratch is highly under-constrained and thus inherently unstable.
To address this, our optimization process for {\name} is divided into three sequential stages, including facial mesh optimization, canonical hair optimization, and joint optimization.
%

\vspace{\bfskip}
\noindent
\textbf{Learnable parameters.} For clarity, we list all learnable parameters here.
For the Gaussian hair, 
$\theta_{gs}$ refers to all learnable parameters of the canonical Gaussian hair, while $\theta_{def}$ represents the MLP parameters for the hair deformation field.
For the facial mesh,
$\hat{\bm{T}}_{di}$ is a learnable latent map (i.e., neural texture~\cite{DBLP:journals/tog/ThiesZN19}) for representing diffuse color.
$\theta_{v}$ refers to the parameters of the view texture decoder, from which a view-dependent latent map $\hat{\bm{T}}_{v}$ is produced.
$\theta_{dy}$ refers to the parameters of the dynamic texture decoder, from which an expression-dependent latent map $\hat{\bm{T}}_{dy}$ is produced.
$\theta_{disp}$ represents the parameters of the geometry decoder, from which a UV displacement map $\hat{\bm{G}}_d$ is produced.
%
$\theta_{pix}$ specifies the parameters of the pixel decoder $\mathcal{D}_{pix}$ that decodes neural textures into RGB colors.

\vspace{\bfskip}
\noindent
\textbf{Optimizing Facial Mesh.} In the first stage, we optimize all learnable parameters related to the facial mesh (i.e., $\hat{\bm{T}}_{di}, \theta_{v}, \theta_{dy}, \theta_{disp}$, and $\theta_{pix}$) with two per-pixel photometric losses $\mathcal{L}^F_{pho}$ and $\mathcal{L}^F_{di\cdot pho}$, a D-SSIM loss $\mathcal{L}^F_{ssim}$, a shrink loss $\mathcal{L}^F_{shr}$, two depth-based losses $\mathcal{L}^F_d$ and $\mathcal{L}^F_n$, and three regularization losses $\mathcal{L}^F_{lap}$, $\mathcal{L}^F_{nc}$, and $\mathcal{L}^F_{el}$.

\noindent\textit{Photometric losses.}
$\mathcal{L}^F_{pho}$ and $\mathcal{L}^F_{ssim}$ provide supervisions for rendered facial colors as:
\begin{equation}
\setlength{\abovedisplayskip}{\eqskip}
\setlength{\belowdisplayskip}{\eqskip}
\begin{split}
    &\mathcal{L}^F_{pho} = || \bm{I}_{head} - \bm{\hat{I}}_{head} ||_2, \\
    &\mathcal{L}^F_{ssim} = 1 - \text{SSIM}(\bm{I}_{head}, \bm{\hat{I}}_{head}),
\end{split}
\end{equation}
where $\bm{I}_{head}$ is the ground truth image of the head part.

We introduce an extra L2-based photometric loss $\mathcal{L}^F_{di\cdot pho} = || \bm{I}_{head} - \bm{\hat{I}}^{di}_{head} ||_2$ to promote more meaningful texture decomposition,
where $\bm{\hat{I}}^{di}_{head}$ is decoded from only the diffuse latent textures $\hat{\bm{T}}_{di}$.

\vspace{\bfskip}
\noindent\textit{Geometric losses.}
We use depth and screen-space normal losses to refine the geometry of the facial mesh as follows:
\begin{equation}
\setlength{\abovedisplayskip}{\eqskip}
\setlength{\belowdisplayskip}{\eqskip}
\begin{split}
    &\mathcal{L}^F_d = || (\bm{D}_{h} - \bm{\hat{D}}_{h}) \odot \bm{M}_d ||_1, \\ &\mathcal{L}^F_n = || N(\bm{D}_{h}) - N(\bm{\hat{D}}_{h}) \odot \bm{M}_d ||,
\end{split}
\end{equation}
where $\bm{D}_{h}$ denotes the depth map derived from multi-view images using Metashape software~\cite{metashape}.
$\bm{\hat{D}}_{h}$ is the depth map rasterized by our facial mesh and $N(\cdot)$ calculates screen space normals~\cite{DBLP:conf/cvpr/MaSSWLTS21}.
$\bm{M}_d$ is used to penalize those pixels whose depth errors are less than a depth threshold $\delta_D$ (set to 5mm), minimizing the effect of noise.

\vspace{\bfskip}
\noindent\textit{Shrink loss.}
To address the issue of the FLAME scalp often being oversized and overlapping with the hair, we introduce a shrink regularization loss $\mathcal{L}^F_{shr}$ for the scalp vertices,
{
\small
\begin{equation}
\setlength{\abovedisplayskip}{\eqskip}
\setlength{\belowdisplayskip}{\eqskip}
    \mathcal{L}^F_{shr} = || \mathcal{V}^{scalp}_r(\beta, \psi, \phi) - \text{Mean}(\mathcal{V}^{scalp}(\beta, \psi, \phi)]) ||_2,
\end{equation}
}
where $\mathcal{V}^{scalp}_r$ are the scalp vertices visible in the current frame and are obtained by projecting hair masks back to the deformed FLAME mesh.
By shrinking the scalp towards a fixed center, the Gaussians can be optimized to their correct locations without being obscured by a wrong scalp mesh.

\vspace{\bfskip}
\noindent\textit{Regularizations.}
Three regularization losses are used to ensure a reasonable facial mesh (e.g., no face crossing, reversing). 
%
%
The mesh Laplacian loss $\mathcal{L}^F_{lap}$ and normal consistency loss $\mathcal{L}^F_{nc}$ smooth the facial mesh, while the edge length loss $\mathcal{L}^F_{el}$ keeps the rigidity of the mesh as much as possible.

In summary, the complete training loss for our facial mesh is formulated as a weighted sum of these loss terms:
\begin{equation}
\setlength{\abovedisplayskip}{\eqskip}
\setlength{\belowdisplayskip}{\eqskip}
    \begin{split}
        \mathcal{L}^F = &\;\lambda_{p} \mathcal{L}^F_{pho} + 3\cdot \lambda_{p} \mathcal{L}^F_{di\cdot pho} + \lambda_d \mathcal{L}^F_d \\
        + &\;\lambda_n \mathcal{L}^F_n + \lambda_{ss} \mathcal{L}^F_{ssim} + \lambda_{sh} \mathcal{L}^F_{shr} + \mathcal{L}^F_{reg},
    \end{split}
\end{equation}
where $\mathcal{L}^F_{reg} = \lambda_{lap} \mathcal{L}^F_{lap} + \lambda_{nc} \mathcal{L}^F_{nc} + \lambda_{el} \mathcal{L}^F_{el}$.

\vspace{\bfskip}
\noindent
\textbf{Optimizing Canonical Gaussian Hair.}
Following the initialization with points sampled around the scalp mesh (as mentioned in Sec.~\ref{sec:canonical_hair}), we optimize the canonical Gaussian hair parameters (i.e., $\theta_{gs}$) using two appearance losses $\mathcal{L}^H_{pho}$ and $\mathcal{L}^H_{ssim}$ as in 3DGS~\cite{DBLP:journals/tog/KerblKLD23}, a silhouette loss $\mathcal{L}^H_{silh}$, and a regularization loss $\mathcal{L}^H_{sol}$.

Specifically, two appearance losses are defined as:
\begin{equation}
\setlength{\abovedisplayskip}{\eqskip}
\setlength{\belowdisplayskip}{\eqskip}
\begin{split}
    &\mathcal{L}^H_{pho} = || \bm{I}_{hair} - \bm{\hat{I}}_{hair} ||_2, \\
    &\mathcal{L}^H_{ssim} = 1 - \text{SSIM}(\bm{I}_{hair}, \bm{\hat{I}}_{hair}),
\end{split}
\end{equation}
where $\bm{I}_{hair}$ is the ground truth image of the hair part.

To encourage better disentanglement between the facial mesh and Gaussian hair, we introduce a silhouette loss:
\begin{equation}
\setlength{\abovedisplayskip}{\eqskip}
\setlength{\belowdisplayskip}{\eqskip}
\begin{split}
    \mathcal{L}^H_{silh} =& || (\bm{M}_{hair} - \bm{\hat{A}}_{hair}) \odot \bm{\Delta} ||_1, \;\;\; \\
    \text{where }& \bm{\Delta}(x_i)  = \underset{x_j \in \bm{M}_{hair}}{min} ( ||x_i - x_j ||_2).
\end{split}
\end{equation}
where $\bm{M}_{hair}$ is the ground truth hair mask, obtained using a standard facial parsing algorithm~\cite{CelebAMask-HQ}.
$\bm{\Delta}(\cdot)$ is a weighting function that ensures distant incorrect pixels in the rendered mask are penalized more heavily than pixels that are closer.
%

We also introduce a regularization loss that encourages the Gaussian hair to generate a solid hair mask, except for its boundary regions.
Mathematically, this loss is defined as:
$\mathcal{L}^c_{sol} = || (\bm{1} - \bm{\hat{A}}_{hair}) \odot \text{Erode}(\bm{M}_{hair}) ||_1$, where $\text{Erode}(\cdot)$ represents the erosion operation.

In summary, the complete loss used to train our canonical hair is defined as:
{
\begin{equation}
\setlength{\abovedisplayskip}{\eqskip}
\setlength{\belowdisplayskip}{\eqskip}
\begin{split}
    \mathcal{L}^H = \lambda_{p} \mathcal{L}^H_{pho} + \lambda_{ss} \mathcal{L}^H_{ssim} + \lambda_{sil} \mathcal{L}^H_{silh} + \lambda_{sol} \mathcal{L}^H_{sol}.
\end{split}
\end{equation}
}

\vspace{\bfskip}
\noindent
\textbf{Joint Optimization.}
With proper initializations of the neural mesh and canonical 3DGS hair, we jointly optimize the hybrid mesh-Gaussian avatar across all frames, with a primary focus on improving the quality of the face-hair overlapping region.
The objective function is defined as:
{
\small
\begin{equation}
\setlength{\abovedisplayskip}{\eqskip}
\setlength{\belowdisplayskip}{\eqskip}
\begin{split}
    \mathcal{L} = &\;\lambda_{p} \mathcal{L}_{pho} + 3\cdot \lambda_{p} \mathcal{L}^F_{di\cdot pho} + \lambda_{ss} \mathcal{L}_{ssim} + \lambda_{sol} \mathcal{L}^H_{sol} \\ 
    +& \lambda_{n} (||\delta r||_2 + ||\delta s||_2 + ||\delta o||_2 + ||\delta c||_2) + \lambda_{a} \mathcal{L}_{aiap}.
\end{split}
\end{equation}
}
In this stage, we optimize for $\theta_{def}, \hat{\bm{T}}_{di}, \theta_v, \text{and } \theta_{dy}$.
%
Additionally, we introduce new regularizations to constrain the per-Gaussian update and an as-isometric-as-possible loss $\mathcal{L}_{aiap}$~\cite{DBLP:conf/cvpr/QianWM0024} to encourage the rigidity of the Gaussian hair.
%

\begin{figure}[t]
\centering
\includegraphics[width=\linewidth]{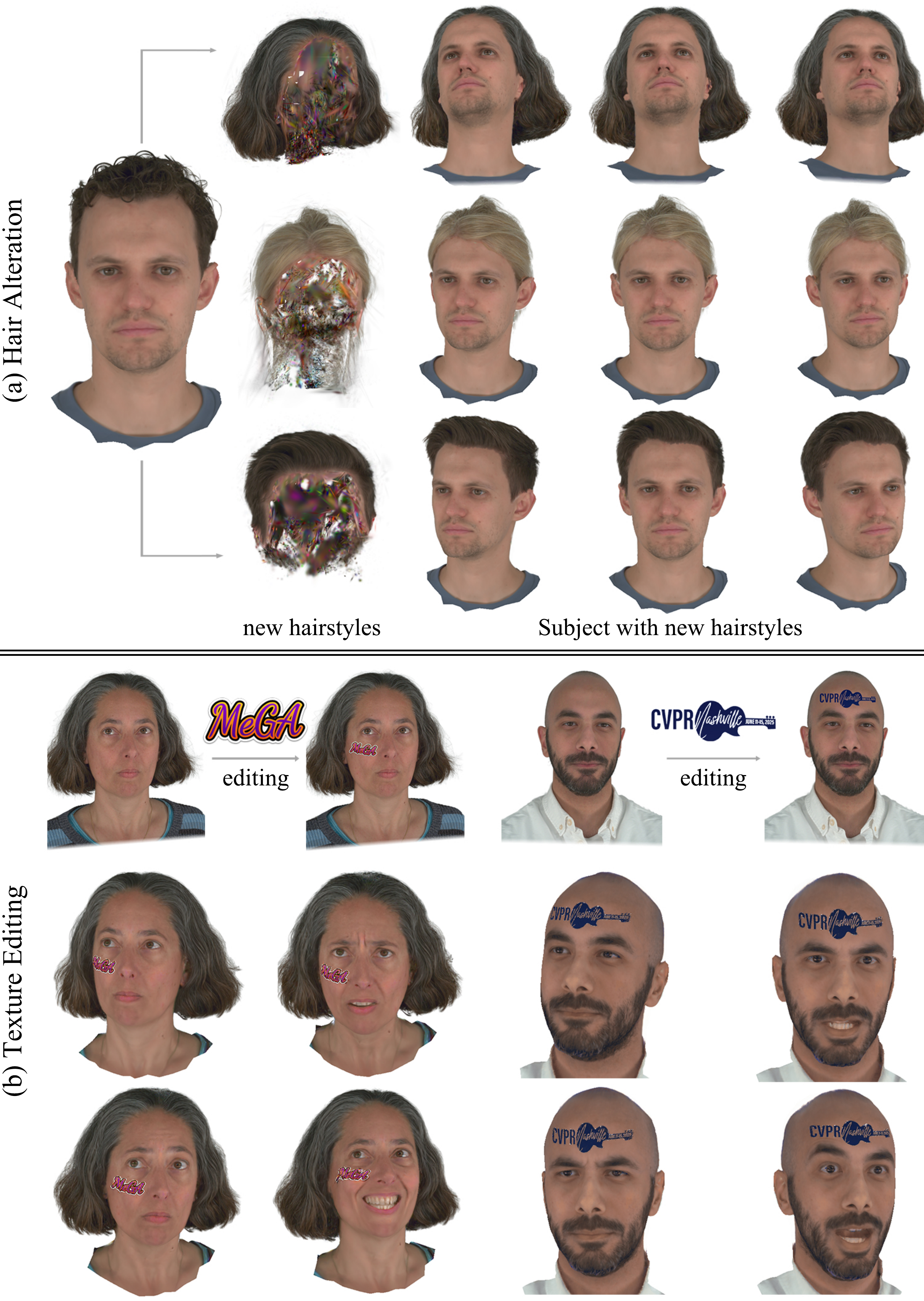}
\caption{\textbf{Hairstyle Alteration and Texture Editing.}
{\name} naturally supports hairstyle alteration and texture editing. 
The edited head avatar can be rendered in novel views and expressions.
}
\label{fig:edit}
\vspace{-2mm}
\end{figure}

\section{Editing Head Avatars}
\label{sec:edit}

Due to the disentangled facial mesh and Gaussian hair, our {\name} naturally facilitates various editing operations.

\vspace{\bfskip}
\noindent
\textbf{Hairstyle Alteration.}
As shown in Fig.~\ref{fig:overview}, our approach can easily update A's hairstyle with B's after alignment (with scaling).
Specifically, we load subject A's facial mesh (i.e., $\hat{\bm{T}}_{di}, \theta_{v}, \theta_{dy}, \theta_{disp}$, and $\theta_{pix}$) and load subject B's Gaussian hair (i.e., $\theta_{gs}$ and $\theta_{def}$).
Then, an ICP-based alignment (with scaling) is conducted to align B's hair to A's.
%

%



\vspace{\bfskip}
\noindent
\textbf{Facial Texture Editing.} 
%
Our \name\ can easily support texture editing by updating the diffuse neural texture map $\hat{\bm{T}}_{di}$ according to the painted image $\bm{I}_p$ and its corresponding mask $\bm{M}_p$ similar to NeuMesh~\cite{DBLP:conf/eccv/YangBZBZCZ22}.
Specifically, to edit facial textures, we first remap the 2d painting mask to the UV space, obtaining a mask $\bm{M}^{uv}_p$.
Only the latent codes under this mask are optimized during the subsequent optimization process.
%
Then we optimize these codes in the diffuse texture map $\hat{\bm{T}}_{di}$ with a learning rate $0.01$ and the pixel decoder $\theta_{pix}$ with a learning rate $0.0001$.
Slightly finetuning the pixel decoder allows it to show new colors that are not seen during training head avatars.

Note that we calculate losses for the complete image on the view $\bm{I}_p$ and calculate losses outside the painting mask on other views.
Optimizing the losses on other views serves as a regularization of the pixel decoder $\mathcal{D}_{pix}$, ensuring minimal changes on the non-painting regions.
%
%




\begin{figure*}[t]
    \centering
    \includegraphics[width=0.95\linewidth]{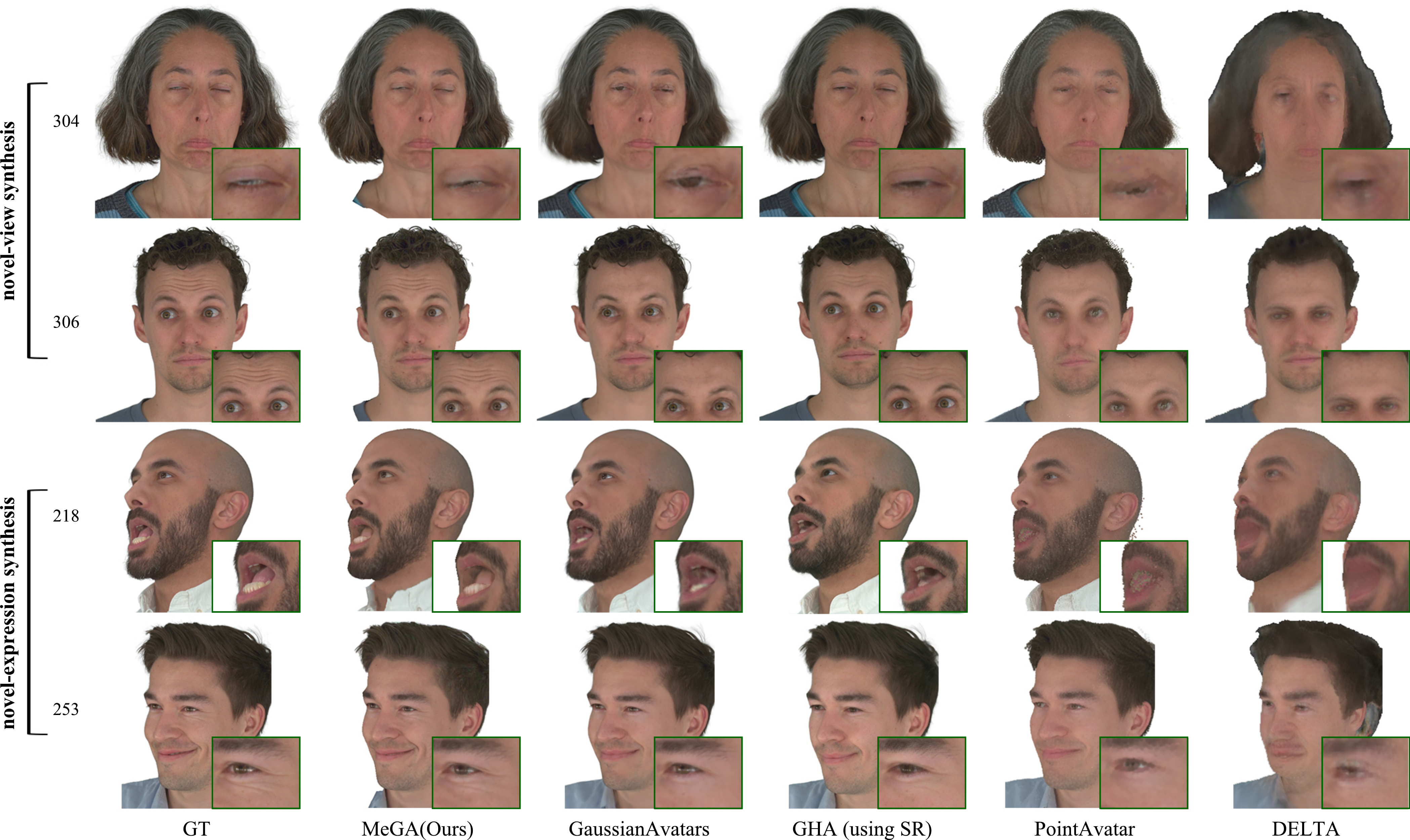}
    \caption{\textbf{Comparisons with State-of-the-Art Methods.}
    {\name} generates more realistic facial renditions compared to previous state-of-the-art methods, especially in terms of detailed skin textures.
    Note that Gaussian Head Avatar (GHA) uses a super-resolution (SR) module.
    }
    \label{fig:sota}
    \vspace{-1mm}
\end{figure*}

\section{Experiments}

We evaluate our approach on the NeRSemble dataset~\cite{DBLP:journals/tog/KirschsteinQGWN23}, which contains multi-view videos of each subject and calibrated camera parameters of all 16 cameras.
GaussianAvatars~\cite{DBLP:conf/cvpr/QianKS0GN24} downsample the images to a resolution of $802 \times 550$ and generate a foreground mask for each image.
Based on their processed images, we further obtain facial parsing results for each image using an open-source algorithm~\cite{CelebAMask-HQ} and depth maps for each frame using Metashape software~\cite{metashape}.

We train our MeGA using the same train/test splits as GaussianAvatars~\cite{DBLP:conf/cvpr/QianKS0GN24}.
Specifically, 9 out of 10 expression sequences and 15 out of 16 available cameras are used for training, while the remaining camera and expression sequence are reserved for evaluation.
%
All metrics are calculated based on image pixels under the rasterization mask.
The facial geometry is evaluated using the Mean Absolute Error (MAE) between the reconstructed depth maps and our rasterized depth maps.

\begin{table}[t]
\centering
\caption{
\textbf{Comparisons with State-of-the-Art Methods.}
{\name} achieves better LPIPS, SSIM, and PSNR (1dB higher than the $2^{\text{nd}}$ best method).
We bold (underline) the best ($2^{\text{nd}}$ best) results.
}
\resizebox{\linewidth}{!}{%
\begin{tabular}{l|ccc|ccc}
\toprule
    \multirow{2}{*}{Method} & \multicolumn{3}{c|}{Novel-View Synthesis} & \multicolumn{3}{c}{Novel-Expr. Synthesis} \\
    & PSNR $\uparrow$ & SSIM $\uparrow$ & LPIPS $\downarrow$ & PSNR $\uparrow$ & SSIM $\uparrow$ & LPIPS $\downarrow$ \\
\midrule
DELTA & 24.62 & 0.871 & 0.138 & 22.60 & 0.858 & 0.158 \\
PointAvatar & 27.08 & 0.918 & 0.091 & 25.79 & 0.916 & 0.103 \\
Gaussian Head Avatar & 29.48 & 0.894 & 0.084 & 22.02 & 0.847 & 0.156 \\
GaussianAvatars & \uline{33.54} & \uline{0.951} & \uline{0.055} & \uline{31.45} & \uline{0.947} & \uline{0.060} \\
\midrule
MeGA(Ours) & \bm{{34.11}} & \bm{{0.954}} & \bm{{0.052}} & \bm{{32.59}} & \bm{{0.949}} & \bm{{0.057}} \\
\bottomrule
\end{tabular}
}
\label{tab:sota}
\end{table}

\subsection{Comparisons with State-of-the-Art Methods}

We conduct comparisons with GaussianAvatars~\cite{DBLP:conf/cvpr/QianKS0GN24}, Gaussian Head Avatar with super-resolution (SR) \cite{DBLP:conf/cvpr/XuCL00ZL24}, PointAvatars~\cite{DBLP:conf/cvpr/ZhengYWBH23}, and DELTA~\cite{DBLP:journals/corr/abs-2309-06441} to demonstrate the superiority of our method.
All baselines are trained from scratch using their public codes and the training details are provided in our \supp{} 
%
%
Tab.~\ref{tab:sota} shows that our approach achieves the best PSNR, SSIM, and LPIPS averaged among all 9 subjects.
As shown in Fig.~\ref{fig:sota}, due to the use of expression-dependent dynamic textures (i.e., $\hat{\bm{T}}_{dy}$), our MeGA can model more subtle geometric details (e.g., wrinkles in subject 306 and 253).
Besides, due to the integral-based rendering, 3DGS-based facial rendering tends to produce blurry or interpenetrated results around the eye and mouth regions (e.g., subject 304 and 218).
The possible reason is that when fitting a close-eye expression, the pixels around the eye region should ideally only use the Gaussians of the eyelids for rendering. 
However, the 3DGS rendering process cannot distinguish between the Gaussians of the eyelids and eyeballs, thereby using both of them to perform rendering and producing interpenetrated artifacts. 
The blurry mouth is caused by a similar reason.
%
%
%
Note that while Gaussian Head Avatar produces promising results for novel-view synthesis, it struggles with novel expression rendering due to its heavy reliance on the implicit deformation and super-resolution module.
More results are shown in our \supp{}
%
%

\subsection{Experiments on Head Editing}

We present our results for qualitative evaluation only, as, to the best of our knowledge, no previous methods\footnote{Previous mesh-based methods~\cite{DBLP:conf/cvpr/MaSSWLTS21, DBLP:conf/cvpr/GrassalPLRNT22, DBLP:journals/corr/abs-2309-06441} are not suitable for texture editing due to the entanglement of base colors and view-dependent effects. DELTA~\cite{DBLP:journals/corr/abs-2309-06441} produces rather poor renderings in our settings.} are suitable for the following types of head editing.


{\name} supports changing someone's hairstyle to a new one (i.e., short, medium, and long hair) from another \name{}-pretrained model (Fig.~\ref{fig:edit}a). 
%
Recomposed head avatars can be rendered in novel views and expressions.
Fig.~\ref{fig:edit}b demonstrates the texture editing functionality.
Given a painted image of the subject and the corresponding painting mask, {\name} can embed this modification into the 3D head avatar and render view-consistent images in novel views and expressions.

\begin{table*}[t]
\centering
\caption{
\textbf{Ablation Studies on Subject 306.}
We demonstrate the effectiveness of each proposed component.
(a) shows the importance of our disentangled texture maps in generating high-quality renderings.
%
(b) shows the positive effects of our mesh subdivision and UV displacement map.
(c) shows the superiority of our occlusion-aware blending.
%
A blank entry indicates the same settings as ``MeGA (Ours)''.
}
\resizebox{0.84\linewidth}{!}{%
\begin{tabular}{c|l|l|c|c|c|c|c|c}
\toprule
\multirow{2}{*}{Label} & \multirow{2}{*}{Name} & \multirow{2}{*}{Texture} & \multirow{2}{*}{Mesh Geom.} & \multirow{2}{*}{Blending} & \multirow{2}{*}{PSNR $\uparrow$} & \multirow{2}{*}{SSIM $\uparrow$} & \multirow{2}{*}{LPIPS $\downarrow$} & \multirow{2}{*}{Geo. MAE $\downarrow$} \\
& & & & &  &  &  & \\
\midrule
& {\name} (Ours) & & & & \bm{{33.57}} & \bm{{0.963}} & \bm{{0.040}} & \bm{{2.25mm}} \\
\midrule
(a.1) & {\name}-noview & w/o $\hat{T}_v$ & & & 31.68 & 0.958 & 0.053 & 2.87mm \\
(a.2) & {\name}-nodyn & w/o $\hat{T}_{dy}$ & & & 32.81 & 0.959 & 0.050 & 2.91mm\\
\midrule
(b.1) & {\name}-nosubdiv & & no subdivision & & 32.81 & 0.962 & 0.046 & 3.38mm \\
(b.2) & {\name}-nodisp & & no $\hat{\bm{G}}_d$ & & 32.99 & 0.959 & 0.054 & 7.48mm \\
\midrule
(c.1) & {\name}-gsdepth & & & using alpha-acc. depths & 27.94 & 0.950 & 0.068 & \bm{{2.25mm}} \\
(c.2) & {\name}-allGS & & & no early-stopping & 31.42 & 0.957 & 0.048 & \bm{{2.25mm}} \\
\bottomrule
\end{tabular}
}
\label{tab:ablation-study}
\end{table*}

\subsection{Ablation Studies}

In this section, we present a series of ablation studies to verify the effectiveness of our major design choices.

\begin{figure}[t]
\centering
\includegraphics[width=\linewidth]{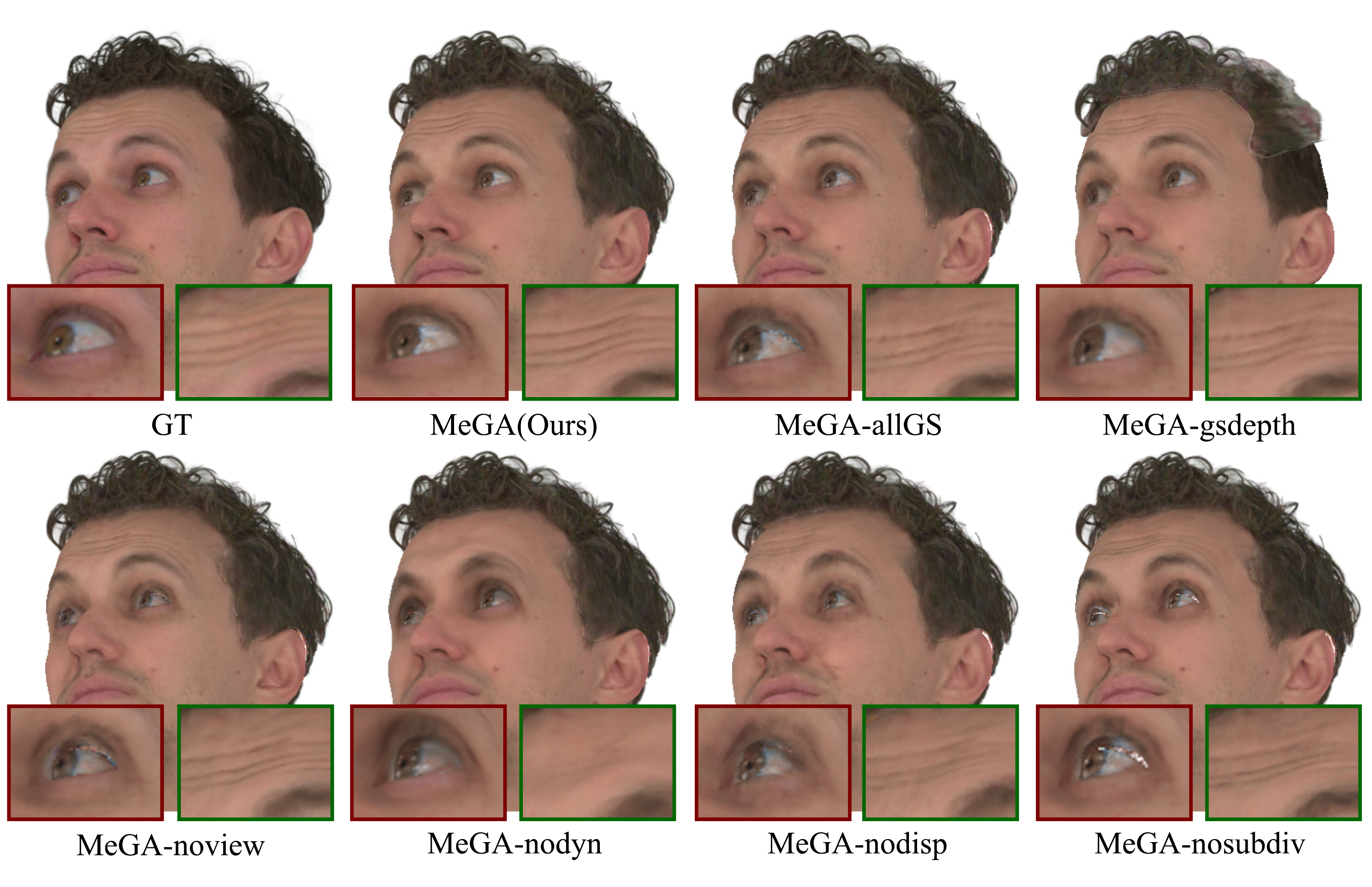}
\caption{\textbf{Ablation Studies} on Disentangles Texture Maps, Geometry Refinement, and Blending Strategies.
Disabling the view texture $\hat{T}_v$ and dynamic texture $\hat{T}_{dy}$ loses the highlights in the eyes and the forehead wrinkles, respectively.
Removing any component in our geometry refinement and occlusion-aware blending module degrades the final renderings.
}
\label{fig:ab_texmap}
\vspace{0mm}
\end{figure}

\vspace{\bfskip}
\noindent
\textbf{Disentangled Texture Maps.}
Tab.~\ref{tab:ablation-study} (a.1) and (a.2) illustrate the roles of our two disentangled texture maps, with the corresponding visual results shown in Fig.~\ref{fig:ab_texmap}.
When the view texture $\hat{\bm{T}}_v$ is disabled (MeGA-noview), {\name} struggles to handle view-dependent effects and fails to capture highlights in the eyes.
%
%
When the expression-dependent dynamic texture $\hat{\bm{T}}_{dy}$ is disabled (MeGA-nodyn), {\name} loses the ability to model detailed skin appearance (e.g., the forehead wrinkles).
Disabling any of them results in worse quantitative metrics (31.68/32.81 vs. 33.57).

\vspace{\bfskip}
\noindent
\textbf{Mesh Geometry.}
We investigate the effect of mesh subdivision and the use of the UV displacement map $\hat{\bm{G}}_d$ for enhancing geometry details.
The quantitative results are reported in Tab.~\ref{tab:ablation-study} (b.1) and (b.2).
Without mesh subdivision, only 5023 vertices are adapted to fit the facial depths, leading to inferior facial geometry and renderings (3.38mm vs. 2.25mm Geo. MAE, 32.81 vs. 33.57 PSNR).
Using a UV displacement map $\hat{\bm{G}}_d$ significantly improves the evaluation metrics (2.25mm vs. 7.48mm Geo. MAE, 33.57 vs 32.99 PSNR).
The visual results are shown in Fig.~\ref{fig:ab_texmap}.

%

\vspace{\bfskip}
\noindent
\textbf{Blending Strategies.}
To verify the effectiveness of our mesh occlusion-aware blending approach, we test alternative blending strategies and report the quantitative results in Tab.~\ref{tab:ablation-study} (c.1)-(c.2).
``{\name}-gsdepth'' attempts to obtain the visibility of the Gaussian hair using 3DGS-rendered depths, instead of the ``near-z'' depths.
However, 3DGS-rendered depths may fluctuate due to minor training errors and make occlusion relations between the head and 3DGS hair changing constantly, resulting in the optimization objective of 3DGS shifting throughout the training process and unstable optimization.
``MeGA-allGS'' disables our early-stopping strategy and uses both invisible and visible Gaussians for hair rendering.
In this case, if a single Gaussian mistakenly appears in front of the facial mesh, the invisible Gaussians will be used to fit the facial appearance,  disrupting the learning of facial textures and leading to inferior facial renderings (31.42 vs. 33.57 PSNR and Fig. \ref{fig:ab_texmap}).
%
%
%
%

\vspace{\bfskip}
\noindent
\textbf{Loss Functions.}
%
%
Removing any loss function degrades the performance.
More details are shown in our \supp{}
%

\section{Conclusion}
In this paper, we present hybrid mesh-Gaussian head avatars (\name{}), which employ neural mesh for face modeling and 3DGS for hair modeling.
For high-quality facial modeling, we enhance the FLAME mesh and decode a UV displacement map for personalized geometric details.
Facial colors are decoded via a lightweight MLP from a neural texture map that consists of disentangled diffuse texture $\hat{\bm{T}}_{di}$, view-dependent texture $\hat{\bm{T}}_v$, and dynamic texture $\hat{\bm{T}}_{dy}$.
For high-quality hair modeling, we build a static 3DGS hair and employ a rigid transformation combined with an MLP-based deformation field for animation.
The final renderings are obtained by blending the hair and head parts with our occlusion-aware blending module.
In addition to achieving the best rendering results, \name{} naturally supports various editing functionalities, including hairstyle alteration and texture editing.

\clearpage

\section*{Acknowledgements}

This work was supported by the National Key Research and Development Program of China (No. 2023YFF0905104), the Natural Science Foundation of China (No. 62361146854) and Tsinghua-Tencent Joint Laboratory for Internet Innovation Technology. 

{
    \small
    \bibliographystyle{ieeenat_fullname}
    \bibliography{ref}
}


\end{document}